%% file: main.tex
\documentclass{article}

\PassOptionsToPackage{numbers, compress}{natbib}

\usepackage[preprint]{main}

\usepackage{subcaption}
\usepackage{subfiles}
\usepackage{tabularx} %
\usepackage{multirow}
\usepackage{hyperref}
\usepackage{amsmath, amssymb, mathrsfs}
\usepackage{cleveref}
\usepackage{graphicx}
\usepackage{wrapfig}  
\usepackage[utf8]{inputenc} %
\usepackage[T1]{fontenc}    %
\usepackage{hyperref}       %
\usepackage{url}            %
\usepackage{booktabs}       %
\usepackage{amsfonts}       %
\usepackage{nicefrac}       %
\usepackage{microtype}      %
\usepackage{xcolor}         %
\usepackage{enumitem}
\usepackage{makecell}
\usepackage{array}
\usepackage{color}
\usepackage[most]{tcolorbox}

\input{math_commands.tex}

\usepackage{xcolor,pifont}
\newcommand{\dataset}[1]{\texttt{#1}}

\newtheorem{definition}{Definition}
\newcommand{\framework}{\textsc{TREK}}
\newcommand{\reasoner}{\textsc{EvoReasoner}}
\newcommand{\updater}{\textsc{EvoKG}}

\title{Temporal Reasoning over Evolving Knowledge Graphs}

\author{%
Junhong Lin$^{1}$, Song Wang$^{2}$, Xiaojie Guo$^{3}$, Julian Shun$^{1}$, Yada Zhu$^{3}$\\
$^1$MIT CSAIL, $^2$University of Virginia, $^3$IBM Research\\
  \texttt{\{junhong,jshun\}@mit.edu} \quad \texttt{sw3wv@virginia.edu} \quad \texttt{\{xiaojie.guo,yzhu\}@ibm.com}
}

\begin{document}

\maketitle

\begin{abstract}
Large language models (LLMs) excel at many language understanding tasks but struggle to reason over knowledge that evolves. To address this, recent work has explored augmenting LLMs with knowledge graphs (KGs) to provide structured, up-to-date information. However, many existing approaches assume a static snapshot of the KG and overlook the temporal dynamics and factual inconsistencies inherent in real-world data.
To address the challenge of reasoning over temporally shifting knowledge, we propose \textbf{\reasoner{}}, a temporal-aware multi-hop reasoning algorithm that performs global-local entity grounding, multi-route decomposition, and temporally grounded scoring. Furthermore, to ensure that the underlying KG remains accurate and up-to-date, we introduce \textbf{\updater{}}, a noise-tolerant KG evolution module that incrementally updates the KG from unstructured documents through confidence-based contradiction resolution and temporal trend tracking. 
We evaluate our approach on temporal QA benchmarks and a novel end-to-end setting where the KG is dynamically updated from raw documents. 
Our method outperforms both prompting-based and KG-enhanced baselines, effectively narrowing the gap between small and large LLMs on dynamic question answering. Notably, an 8B-parameter model using our approach matches the performance of a 671B model prompted seven months later. These results highlight the importance of combining temporal reasoning with KG evolution for robust and up-to-date LLM performance.
Our code is publicly available at \href{https://github.com/junhongmit/TREK}{github.com/junhongmit/TREK}.
\end{abstract}

\section{Introduction}
\label{sec:introduction}
\subfile{sections/1_introduction}

\section{Preliminaries} 
\label{sec:prelims}
\subfile{sections/2_prelims}

\section{Multi-hop Temporal Reasoning over Evolving Knowledge Graphs}
\label{sec:reasoning}
\subfile{sections/3_reasoning}

\section{Noise-Tolerant Knowledge Graph Evolution}
\label{sec:update}
\subfile{sections/4_update}

\vspace{-0.25em}
\section{Experiments}
\label{sec:experiments}
\subfile{sections/5_experiments}

\section{Related Work}
\label{sec:related_works}
\subfile{sections/6_related_works}

\section{Conclusion}
\label{sec:conclusion}
\subfile{sections/7_conclusion}

{
\small
\bibliographystyle{ACM-Reference-Format}
\bibliography{reference}
}

\appendix
\subfile{sections/appendix}

\end{document}

%% file: math_commands.tex
\usepackage{amsmath,amsfonts,bm}

\def\eqref#1{equation~\ref{#1}}

\def\1{\bm{1}}

\def\vs{{\bm{s}}}

\def\vv{{\bm{v}}}

\DeclareMathAlphabet{\mathsfit}{\encodingdefault}{\sfdefault}{m}{sl}
\SetMathAlphabet{\mathsfit}{bold}{\encodingdefault}{\sfdefault}{bx}{n}

\def\gA{{\mathcal{A}}}

\def\gC{{\mathcal{C}}}

\def\gE{{\mathcal{E}}}
\def\gF{{\mathcal{F}}}
\def\gG{{\mathcal{G}}}

\def\gM{{\mathcal{M}}}

\def\gP{{\mathcal{P}}}

\def\gR{{\mathcal{R}}}

\def\gT{{\mathcal{T}}}

\def\gV{{\mathcal{V}}}

\newcommand{\R}{\mathbb{R}}

%% file: sections/1_introduction.tex
Recent years have seen growing interest in combining large language models (LLMs) with knowledge graphs (KGs) to improve factual reasoning and knowledge coverage~\citep{pan2024unifying, kau2024combining}. Generally, this line of research aims to address a core limitation of LLMs: they are trained on static corpora and struggle to adapt to emerging knowledge efficiently. By interacting with structured KGs and utilizing their knowledge, LLMs can better handle complex reasoning tasks and answer queries involving up-to-date or structured knowledge~\citep{wang2024knowledge, ji2023survey}.

\begin{wrapfigure}{r}{0.4\columnwidth}
    \centering
    \includegraphics[width=0.4\columnwidth]{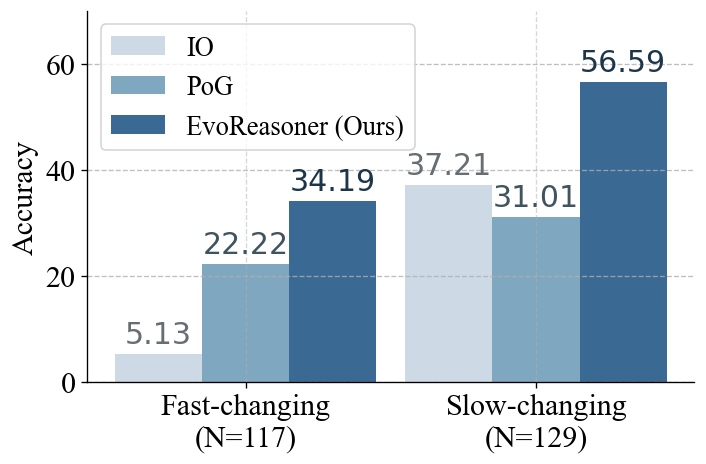}
    \caption{Question answering performance grouped by the required knowledge types in the benchmark \citep{yang2024crag}.}
    \label{fig:examples}
    \vspace{-1em}
\end{wrapfigure}

Despite these advances, many existing KG-augmented approaches~\citep{sunthink, chenplan, luoreasoning, wang2025reasoning} rely on a \textit{static snapshot of the KG} (e.g., downloaded dump of Freebase \citep{bollacker2008freebase} and Wikidata \citep{vrandevcic2014wikidata}), overlooking the \textit{temporal dynamics} of evolving knowledge and factual noise inherent in real-world knowledge. In many cases, temporal metadata is treated as auxiliary information rather than a core signal for inference~\citep{chen2022rotateqvs, yang2024tensor}. 
Our analysis in \Cref{fig:examples} illustrates the limitations of reasoning over static knowledge graphs in scenarios involving temporally evolving or shifting information. This is demonstrated using a question answering (QA) benchmark~\citep{yang2024crag}, where most queries require reasoning over either slowly or rapidly changing facts. As further shown in \Cref{fig:examples},
(1) For questions that require \textbf{slow-changing facts}, the prompting-based method (e.g., IO Prompt) performs well, as internal model knowledge is often sufficient, 
and the inclusion of noisy or contradictory facts from the evolving KG can degrade performance in static-graph reasoning methods like Plan-on-Graph (PoG)~\citep{chenplan}.
(2) For questions that require \textbf{fast-changing facts}, LLMs struggle due to outdated internal knowledge, while graph-based reasoning methods benefit from explicit KG updates. However, these models offer only limited gains, as they lack temporal awareness and are not designed to reason over evolving knowledge.

These limitations stem from two challenges underlying knowledge evolution: \textbf{1. Contradiction and Noise in Static Knowledge.}
KG facts subject to \textit{exclusivity} constraints (e.g., birth date, primary affiliation) can be contradicted by newly added or conflicting information during updates. This necessitates \textit{robust contradiction resolution} strategies that account for context, source reliability, and temporal cues, rather than treating all assertions as equally valid.
\textbf{2. Evolving Trends in Temporal Knowledge.}
\textit{Non-exclusive} facts (e.g, jobs, events) naturally evolve over time, yet existing methods frequently overlook temporal ordering, limiting their effectiveness on time-sensitive queries.

Nevertheless, addressing these challenges demands more than improved reasoning: it requires a unified solution that \emph{jointly} performs temporal reasoning and maintains a consistent, evolving knowledge graph. We propose 
a unified framework that integrates a multi-hop temporal reasoning algorithm with a temporal-aware, noise-tolerant KG evolution mechanism (illustrated in \Cref{fig:overview}). Unlike prior works that treat KG retrieval as a static augmentation layer, our approach tightly couples temporal reasoning with dynamic KG construction and updating. It consists of two key components: (i) \textbf{\reasoner{}}, a temporal multi-hop reasoning algorithm that performs multi-route decomposition, global-local entity grounding, and temporal-aware scoring, and (ii) \textbf{\updater{}}, a noise-tolerant KG evolution method that constructs and updates the graph from unstructured documents. \updater{} applies confidence-based contradiction resolution and temporal trend tracking to maintain robustness and consistency over time.

In experiments, we evaluate \reasoner{} on temporal question answering (QA) benchmarks and a novel end-to-end setting where \updater{} dynamically updates the KG from raw documents. By integrating missing or updated facts, our framework significantly enhances LLM reasoning, outperforming state-of-the-art KG and LLM-only methods with gains of up to \textbf{23.3\%} in temporal reasoning and \textbf{8\%} in evolving KG tasks. Notably, our approach \textbf{closes the gap between small and large models}. For example, a compact LLaMA 3.1–8B \citep{grattafiori2024llama} model, trained in December 2023 and run on a single consumer GPU, improves from \textbf{18.6} to \textbf{37.0\%} after KG updates, comparable to directly prompting a much larger 671B DeepSeek-V3~\citep{liu2024deepseek} model (38.3\%) trained seven months later. This further highlights the effectiveness of our proposed \updater{} framework in improving the reasoning accuracy of LLMs.

Our contributions can be summarized as follows:
\begin{itemize}[topsep=1pt, leftmargin=15pt, itemsep=1pt, parsep=1pt]
\item \textbf{Temporal reasoning method.} 
We introduce \reasoner{}, the first method that combines multi-route decomposition, context-informed global search, and temporal-aware local exploration to enhance LLM reasoning over evolving knowledge graphs. Unlike prior work, \reasoner{} enables precise temporal grounding and robust multi-hop inference under ambiguous or time-sensitive conditions, leading to significantly improved reasoning accuracy.

\item \textbf{Temporal KG evolution method.} 
We propose \updater{}, a KG construction and update method that resolves factual contradictions and models the temporal progression of non-static facts. This enables the KG to remain accurate and consistent over time, supporting reliable reasoning in dynamic real-world scenarios.

\item \textbf{Evaluation across static and temporal dimensions.} 
We evaluate our method on temporal QA benchmarks and a novel end-to-end setting where KGs are updated from raw documents, allowing us to assess the reasoning capability in a temporal setting and how knowledge evolution contributes to downstream QA performance. Our method consistently outperforms both LLM-only and KG-based reasoning baselines across all settings.

\end{itemize}

%% file: sections/2_prelims.tex
\begin{definition}[Attributed Knowledge Graph]
An \emph{attributed knowledge graph} is a directed graph $\gG = (\gV, \gE, \gA, \gR, \gP)$, where each node $v \in \gV$ and each edge $e \in \gE$ has a type given by $\tau(v): \gV \rightarrow \gA$ and $\phi(e): \gE \rightarrow \gR$, respectively. In addition, each node and edge is associated with a property map $\pi: \gV \cup \gE \rightarrow \gP$, where $\gP$ denotes the space of key-value pairs. These property maps allow nodes and edges to store arbitrary attributes in the form of a hashmap.
\end{definition}

\textbf{Problem Formulation.} Let $\gG_t = (\gV_t, \gE_t, \pi_t)$ denote the KG at time $t$, where $\pi_t$ stores attribute metadata for nodes and edges, including temporal information. Each edge $e \in \gE_t$ may be associated with a temporal validity interval $[\tau_{\text{start}}(e), \tau_{\text{end}}(e)] \subseteq \R \cup \{\bot\}$, where $\bot$ denotes an unknown or open-ended temporal boundary, indicating when the associated fact holds true. These start and end times are stored in the property map under explicit keys:
\[
\tau_{\text{start}}(e) = \pi_t(e)[\texttt{valid\ from}], \quad \tau_{\text{end}}(e) = \pi_t(e)[\texttt{valid\ until}]
\]
Given a natural language question $x$, potentially with an explicit or implicit temporal scope (e.g., an explicit temporal question could be: “Who was the president of the U.S. in 2008?”, while an implicit temporal question could be: “Who was the president when the financial crisis happened?”), the goal is to identify a set of entities $\gA \subseteq \gV_t$ by reasoning over the temporally relevant subgraph of $\gG_t$. These entities are then used by the LLM to formulate the final response to the user query.

%% file: sections/3_reasoning.tex
\begin{figure}[t]
    \centering
    \vspace{-1em}
    \includegraphics[width=\columnwidth]{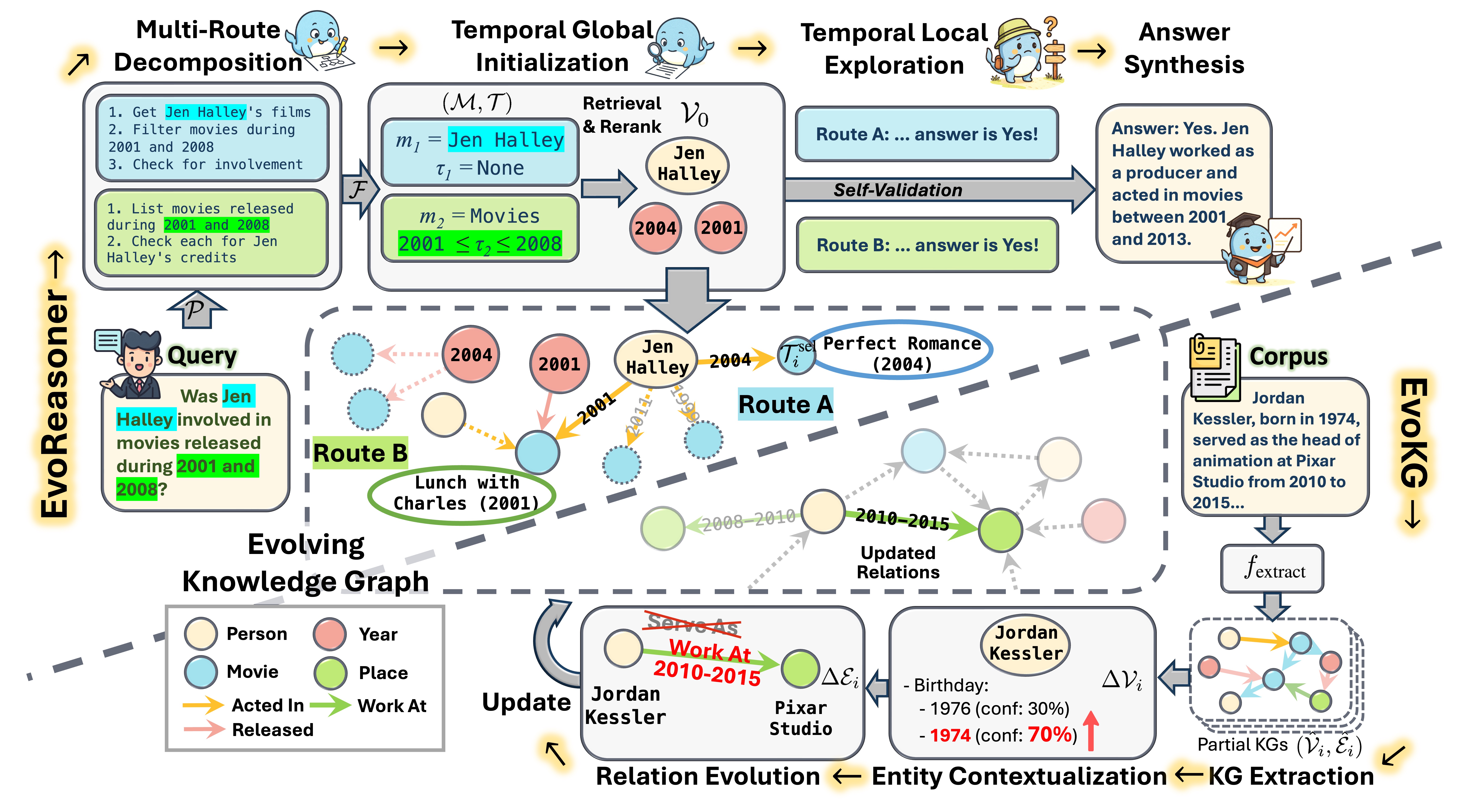}
    \vspace{-1.5em}
    \caption{The overall framework, consisting of two components: \reasoner{} for multi-hop temporal reasoning and \updater{} for document-driven KG evolution. The top pipeline shows how a query is decomposed, grounded to temporal entities, explored locally, and synthesized into an answer. The bottom pipeline illustrates how external documents update the KG via entity and relation contextualization with confidence-aware resolution and temporal evolution.
    }
    \label{fig:overview}
    \vspace{-0.5em}
\end{figure}

We introduce a novel three-stage inference framework, \reasoner{}, as illustrated in \Cref{fig:overview}: (1) \textbf{multi-route decomposition} for diverse and robust exploration; (2) \textbf{global initialization} for temporal-aware query grounding; and (3) \textbf{local exploration} with temporal information. These stages yield multiple reasoning paths, which are aggregated to produce the final answer.

\subsection{Multi-Route Decomposition}

We first decompose the original query $x$ into multiple semantic reasoning routes (e.g., Route A and B in \Cref{fig:overview}), each representing a distinct interpretation or plan for answering the question. This decomposition improves both robustness and recall by accounting for the structural diversity and noise often present in evolving KGs. The resulting routes serve two purposes: (1) they provide \textit{\textbf{soft reasoning guidance}} to direct graph traversal, and (2) offer a \textit{\textbf{self-validation}} mechanism—if multiple routes converge to the same answer, the result is more likely to be reliable. As different routes vary in reasoning efficiency (e.g., number of hops or breadth of search), we prioritize concise and targeted ones, starting from the most efficient and expanding only when necessary to establish consensus.

Let $\gR = \{ r_1, \ldots, r_J \}$ denote the set of candidate decomposition routes, where each route $r_j$ consists of a sequence of subgoals:
$r_j = \{x^{(j)}_1, x^{(j)}_2, \dots, x^{(j)}_{T_j}\}$, where $T_j$ is the length of route $r_j$. 
We define a route planning function $\gP$ (see \Cref{sec:prompt_template} for prompt template) that maps the query to these candidate plans:
$x \xrightarrow{\gP} \gR$.
To evaluate the cost of each route, we propose a multi-objective cost function:
\[
\text{Cost}(r_j) = \sum_{t=1}^{T_j} \psi(x_t^{(j)}), \ \text{where}\ \psi(x_t^{(j)}) = (b_t \cdot n_t)^{h_t}.
\]
$\psi(x_t^{(j)})$ reflects traversal complexity of subgoal $x_t^{(j)}$, based on: $b_t$, the number of candidate relation types from the current entity (branching factor); $n_t$, the number of entities reachable through those relations; and $h_t$, the estimated number of hops required to complete the subgoal.

This formulation encourages the selection of reasoning plans that are narrower and shorter, reducing the chance of semantic drift during exploration. To balance efficiency and diversity, we retain the top-$N$ lowest-cost and semantically distinct plans:
\[
\gR^* = \operatorname{TopK}_N \left( \left\{ r_j \in \gR : -\text{Cost}(r_j) \right\} \right).
\]
These selected routes are used to guide the traversal process, offering complementary reasoning strategies that can be aggregated for answer synthesis and justification.

\subsection{Temporal Contextualized Global Initialization} \label{sec:global_init}
Given a soft reasoning guidance, our method begins by identifying topic entities in $\gG_t$ that are semantically and temporally aligned with the query $x$ and its subgoals $\{ x_t^{(j)} \}$. Prior works (e.g.,~\citet{yang2024crag, sunthink}) typically extract topic entities directly using simple string matching, without verifying their presence in the KG or accounting for the temporal context. While this may suffice in static and clean KGs, it fails in evolving and noisy KGs. For example, a query mentioning “Pixar Company” cannot be grounded if the KG only contains the entity “Pixar Animation Studio.” Similarly, “Oscar Awards” lacks temporal specificity unless linked to one of its 100 historical variants.

To address these challenges, we propose a \textit{context-aware global initialization} strategy that leverages both semantic similarity and temporal signal. This ensures robust grounding to valid entities—even in cases of naming ambiguity or temporally-indexed duplicates—and avoids the common failure case where reasoning begins from non-existent or irrelevant nodes. Let $\gF$ represent a query analysis function that produces two aligned outputs, where $\gM$ is a disjoint set of entity mentions in the query and $\gT$ is the corresponding set of temporal contexts (e.g., “between 2001 and 2008” in \Cref{fig:overview}):
\[
\gF(x) \rightarrow \left( \gM, \gT \right), \quad \gM = \langle m_1, \dots, m_n \rangle, \quad \gT = \langle \tau_1, \dots, \tau_n \rangle.
\]
Each pair $(m_i, \tau_i)$ is jointly embedded into a context-aware vector representation: $\vv_{m_i}^{\tau} = \text{Enc}(m_i, \tau_i)$, where $\text{Enc}(\cdot, \cdot)$ is a pre-trained transformer encoder that incorporates both textual and temporal inputs (e.g., sentence-transformers with timestamp-augmented prompts, see \Cref{sec:prompt_template}).
This embedding is then matched against all KG entity embeddings $\{ \vv_e \mid e \in \gV_t \}$. To retrieve grounding candidates from the KG, we use cosine similarity to select the top-$k$ most similar nodes:
\[
\gC_i = \operatorname{TopK}_k \left( \left\{ e \in \gV_t \ :\ \text{sim}(\vv_{m_i}^\tau, \vv_e) \right\} \right),
\]
where $\operatorname{TopK}_k(\cdot)$ returns the $k$ highest-scoring entities under cosine similarity $\text{sim}(\cdot, \cdot)$, and $k$ is a predefined hyperparameter (e.g., $k = 5$).

Each retrieved candidate \( e_j \in \gC_i \) is further verified using a relevance scoring function (see \Cref{sec:prompt_template} for the prompt template), $s_{\text{init}}(e_j \mid x, \tau_i) \in [0, 1]$, which estimates the likelihood that a KG entity $e_j$ is the correct grounding for $m_i$, conditioned on the full question $x$ and its associated temporal context $\tau_i$. This allows for disambiguation among semantically similar or temporally conflicting entities. Let $k'$ be the number of final entities retained per mention. The final set of anchor entities is selected as
\[
\gV_0 = \bigcup_i \operatorname{TopK}_{k'} \left( \left\{ e_j \in \gC_i : s_{\text{init}}(e_j \mid x, \tau_i) \right\} \right).
\]

\subsection{Temporal-Aware Local Exploration}
After global initialization, our approach performs beam search-based local exploration to traverse the KG. At each step, neighboring relations and entities are scored for relevance to the query. Prior methods often overlook the temporal context, which is critical in domains like sports or politics where entities are linked to numerous time-stamped events. Ignoring temporal signals can lead to irrelevant exploration or reliance on random sampling \citep{sunthink}, missing key evidence.

We address this by introducing a temporal-aware strategy that reranks neighbors based on both semantic and temporal alignment with the current subgoal. This enables more focused and accurate multi-hop traversal through time-relevant paths. In particular, at step $i$ of a reasoning route $r_j$, given a current entity $v_t \in \gV_i$, we iteratively perform the following three actions:

\textbf{1. Relation Selection:} We retrieve all outgoing relations $r \in \gR_i(v_t)$ and the relevance scoring function $s_{\text{rel}}(r \mid x, \{ x_t^{(k)} \}_{t=1}^{T_k})$ scores them for relevance to the current query and subgoals.
The top-ranked relations are selected as follows:
$\gR_i^{\text{sel}} = \operatorname{TopK}_k \left( \left\{ r \in \gR_i(v_i) : s_{\text{rel}}(r \mid x, \{ x_t^{(k)} \}) \right\} \right).$

\textbf{2. Entity Expansion:} For each relation $r \in \gR_i^{\text{sel}}$, we retrieve connected triplets of the form $(v_i, r, v’) \in \mathcal{E}_i$, forming a set of candidate triplets:
$\gT_i = \{ (v_i, r, v’) \in \gE_i : r \in \gR_i^{\text{sel}} \}.$

\textbf{3. Temporal-Aware Reranking:} We perform semantic reranking of triplets using temporal and contextual signals. Each triplet $(v_i, r, v’) \in \gT_i$ is first verbalized into a natural language string (e.g., “Team A played against Team B in 2020”) incorporating both its semantic content and temporal validity interval $[\tau_{\text{start}}, \tau_{\text{end}}] \subseteq \mathbb{R} \cup \{\bot\}$, where $\bot$ denotes an unknown time. Then, the temporal-aware similarity score is computed as  $s_{\text{triplet}}(x, \{x_i\}_{i=1}^{T_k}, (v_t, r, v’)) = \text{sim} \left( x, (v_t, r, v’)) \right)$, i.e., the similarity between embeddings of $x$ and the verbalized triplet. We select the top-ranked triplets as follows:
\[
\gT_i^{\text{sel}} = \operatorname{TopK}_k \left( \left\{ t \in \gT_i : s_{\text{triplet}}(x, \{ x_t^{(k)} \} \mid t) \right\} \right), \ \gV_{i+1} = \{ v’ \mid (v_i, r, v’) \in \gT_i^{\text{sel}} \}.
\]
The destination entities in $\gV_{i+1}$ from the selected triplets define the next search frontier.
The reasoning proceeds until either (1) a stopping criterion is met (i.e., the answer is found or the maximum depth is reached), or (2) no high-confidence edges are available for expansion.

\subsection{Answer Synthesis and Justification} Once valid paths are collected, we synthesize a final answer by aggregating high-confidence paths. Each path $\gP = \{ (r_1, v_1), \dots, (r_L, v_L) \}$, starting from anchor $v_0$, is assigned a confidence score:
\[
\text{Conf}(\gP) = s_{\text{init}}(v_0) \cdot \prod_{i=1}^{L} s_{\text{rel}}(r_i) \cdot s_{\text{triplet}}(v_{i-1}, r_i, v_i),
\]
where $s_{\text{init}}(v_0)$ is the global initialization score from \Cref{sec:global_init}, $s_{\text{rel}}(r_i)$ scores relation relevance at step $i$, and $s_{\text{triplet}}$ captures the temporal and semantic alignment of the traversed edge. For interpretability, we optionally visualize these reasoning paths as annotated subgraphs.

\textbf{Majority Voting Across Reasoning Routes.}
Let \( \gR^* = \{ r_1, \dots, r_N \} \) denote the top-ranked reasoning routes from the multi-route decomposition, each yielding a predicted answer \( a_i \in \gV_t \) with confidence score \( \text{Conf}(\gP_i) \). The final answer is selected via weighted voting:
\[
a^* = \arg\max_{a \in \mathcal{A}} \sum_{i : a_i = a} \text{Conf}(\gP_i).
\]
With confidence scores, we improve robustness by consolidating predictions from multiple plausible reasoning paths, reducing the impact of individual errors or noisy subgoals.

%% file: sections/4_update.tex
We present \updater{}, a temporal-aware framework for constructing and updating knowledge graphs from unstructured text. Our proposed approach aims to resolve following two challenges:

\textbf{1. Contradiction and Noise in Static Knowledge.}
Certain relations (e.g., birth date and primary affiliation) are \textit{exclusive}: only one value should hold at a time. Let $\gR_{\text{excl}} \subset \gR$ denote the set of such relations. Updates to these facts can introduce contradictions, especially when sources are noisy or inconsistent. We model exclusive relations as entity-level properties via the map $\pi: \gV \to \gP$, where each relation $r \in \gR_{\text{excl}}$ is associated with a set of candidate values:
\[
\pi(v)[r] = \left\{ (o_1, \text{conf}_1, \text{ctx}_1), \ldots, (o_k, \text{conf}_k, \text{ctx}_k) \right\},
\]
where $o_i$ is a value candidate, $\text{conf}_i \in [0,1]$ reflects confidence (based on frequency, recency, and source reliability), and $\text{ctx}_i \in \gC$ captures the extraction context. Unlike prior approaches~\citep{lu2025knowledge} that overwrite conflicting values, \updater{} retains multiple candidates with associated confidence, enabling context-sensitive reasoning and improving robustness to noise.

\textbf{2. Evolving Trends in Temporal Knowledge.}
Non-exclusive relations (e.g., works as and acted in) naturally change over time. Let $\gR_{\text{non-excl}} = \gR \setminus \gR_{\text{excl}}$ denote this set of temporally evolving relations. For $r \in \mathcal{R}_{\text{non-excl}}$, we model temporal dynamics via edges:
\[
e = (v_s, r, v_t, [t_{\text{from}}, t_{\text{to}}]) \in \mathcal{E},
\]
where $t_{\text{from}}, t_{\text{to}} \in \mathbb{R} \cup \{\bot\}$ represent the validity interval.
The interval is stored in the property map:
\[
\phi(e) \in \mathcal{R}_{\text{non-excl}}, \quad \pi(e)[\texttt{valid\ from}] = \tau_{\text{start}}, \quad \pi(e)[\texttt{valid\ until}] = \tau_{\text{end}}.
\]
This temporal representation allows \updater{} to preserve the full evolution of entity relations, rather than collapsing them into static facts. For instance, the entity Obama may simultaneously hold (Obama, \texttt{works as}, Senator, [2004, 2009]) and (Obama, \texttt{works as}, President, [2009, 2017]).

To address these two challenges, we develop a two-stage evolution framework: (1) \textbf{entity contextualization} to address static contradiction and ambiguity in extracted nodes; (2) \textbf{relation evolution} to model time-sensitive relational changes.

As shown in \Cref{fig:overview}, given an input document corpus $\mathcal{D} = \{d_1, \dots, d_N\}$, we preprocess it into text chunks $d_i$ and apply a KG extraction function $f_{\text{extract}}$ to produce a partial KG. The resulting triples are then aligned and merged with the current KG, $\gG_t = (\gV_t, \gE_t, \pi_t)$, via a merge function $f_{\text{merge}}$ consisting of the entity contextualization and relation evolution, and produces
\[
d_i \xrightarrow{f_{\text{extract}}} \hat{\gG}_i = (\hat{\gV}_i, \hat{\gE}_i) \xrightarrow{f_{\text{merge}}} (\Delta \gV_i, \Delta \gE_i) \xrightarrow{\text{insert}} \gG_{t+1},
\]
where $\Delta \mathcal{V}_i$ and $\Delta \mathcal{E}_i$ are the sets of new or updated nodes and edges, respectively. This process may be repeated across document batches for comprehensive coverage.

\subsection{Entity Contextualization}\label{sec:entity_alignment}
Document-extracted entities often exhibit lexical ambiguity (e.g., “Pixar Company” vs.\ “Pixar Animation Studio”) or naming collisions (e.g., movies with the same title across years). To maintain KG quality and ensure effective downstream reasoning, we perform a two-stage process:

\textbf{1. Contextual Alignment:} Given extracted entities $\hat{\gV} = \{ \hat{v}_1, \dots, \hat{v}_n \}$, each with name, description, and property map $\pi(\hat{v}_i)$, we aim to align $\hat{v}_i$ to a node $v \in \gV$ or insert it as a new node.
Each candidate is jointly encoded over its type, name, and description, and matched against existing KG nodes:
\[
\vv_i = \text{Enc}(\tau(\hat{v}_i), \text{name}(\hat{v}_i), \text{desc}(\hat{v}_i)),\ \mathcal{A}_i = \operatorname{TopK}_k\left( \{ v \in \mathcal{V} \mid \text{sim}(\vv_i, \vv) \} \right).
\]

Each $v’ \in \gA_i $ is then reranked by a contextual scoring function $s(\hat{v}_i, v’) \in [0,1]$, and the best match is selected: $v^* = \arg\max_{v’ \in \gA(\hat{v})} s(\hat{v}_i, v’)$. If $s(\hat{v}_i, v^*) < \theta$, then $\hat{v}_i$ is inserted as a new node.

\textbf{2. Contextual Merging:} Suppose $v^*$ is the aligned node and $\hat{p} = (r, o, \text{ctx})$ is a candidate property to be merged. For exclusive relations $r \in \mathcal{R}_{\text{excl}}$, the updated property set is
\[
\pi(v^*)[r] \gets \pi(v^*)[r] \cup \{ (o, C(o), \text{ctx}) \},\ C(o) = \frac{\delta\cdot f(o)}{1 + e^{-\gamma \cdot \Delta t(o)}} + (1 - \delta) \cdot w(o),
\]
where $C(o) \in [0,1]$ is the confidence score, with $f(o)$ as the normalized frequency of the value $o$ among prior extractions, $\Delta t(o)$ as the time elapsed since $o$ was last observed (temporal decay), $w(o) \in (0,1]$ as the source credibility weight, and $\gamma$ and $\delta$ as weighting hyper-parameters. Each value $o$ is associated with a context $\text{ctx} \in \mathcal{C}$, e.g., document span, region, or temporal window. These values are preserved in the KG and may later be aggregated or filtered via \textit{Context-Aware Property Grouping}.

Intuitively, in the example of the person “Jordan Kessler” (\Cref{fig:overview}), if the correct birthday 1974 is seen in 7 sources and an incorrect value of 1976 is seen three times, the former achieves a higher frequency and confidence score, ensuring robustness to noise.

\subsection{Relation Evolution}
Given a set of extracted candidate relations $\hat{\mathcal{E}} = \{ (\hat{v}_s, \hat{r}, \hat{v}_t) \}$, the goal is to align or insert each relation into the KG, based on relation schema $\{(\tau(\hat{v}_s), \phi(\hat{r}), \tau(\hat{v}_t))\}$ and semantics. To perform relation evolution, we propose a two-stage process:

\textbf{1. Synonyms Matching:} We first normalize relation types by matching relation schemas. Each candidate relation schema $\hat{s} = (\tau(\hat{v}_s), \phi(\hat{r}), \tau(\hat{v}_t))$ is embedded as $\hat{\vs} = \text{Enc}(\hat{s})$. We compare it against KG schema embeddings $\{ \vs_r \}_{r \in \gR}$ and select the best match:
\[
r^* = \arg\max_{r' \in \mathcal{R}} \text{sim}(\hat{\vs}, \vs_{r'}).
\]
We treat $\phi(\hat{r}) \equiv \phi(r^*)$ as a synonym match when $\max s(\hat{\mathbf{s}}, \mathbf{s}_{r^*}) > \theta$, where $\theta$ is a preset threshold.

\textbf{2. Contextual Alignment and Temporal Merging:} 
Assuming the entity alignments $\hat{v}_s \rightarrow v_s$ and $\hat{v}_t \rightarrow v_t$ (from \Cref{sec:entity_alignment}) are complete, we form the candidate edge: $e = (v_s, r^*, v_t, [\tau_{\text{start}}, \tau_{\text{end}}])$, where the temporal range is derived from the context of extraction.

For non-exclusive relations $r^* \in \gR_{\text{non-excl}}$, we insert the edge directly, preserving all temporally distinct facts. For exclusive relations $r^* \in \gR_{\text{excl}}$, we apply the same confidence-based contradiction handling as in entity merging (see \Cref{sec:entity_alignment}). More details can be found in \Cref{sec:update_details}.

%% file: sections/5_experiments.tex
\newcommand{\res}[2]{#1\textsubscript{$\pm$#2}}
\newcommand{\fir}[2]{\textbf{#1\textsubscript{$\pm$#2}}}
\newcommand{\sed}[2]{\underline{#1 \textsubscript{$\pm$ #2}}}
\newcommand{\thi}[2]{{#1 \textsubscript{$\pm$ #2}}}
\newcommand{\firi}[1]{\textbf{#1}}
\newcommand{\seci}[1]{\underline{#1}}
\newcommand{\thii}[1]{{#1}}
\newcommand{\colorres}[3]{\textcolor{#1}{\textbf{#2} \textsubscript{$\pm$ \textbf{#3}}}}

We evaluate \framework{} on temporal QA benchmarks and a novel end-to-end setting where the KG is dynamically updated from raw documents by \updater{}. Our experiments are designed to answer two key questions:
\textbf{Q1}: \textit{Can \reasoner{} perform accurate and robust multi-hop reasoning over a temporal knowledge graph?}
\textbf{Q2}: \textit{Can \updater{} improve question answering accuracy by evolving an incomplete or outdated KG with new information from raw documents?} 

\textbf{Datasets.} (1) \textbf{KGQA:} To assess multi-hop temporal-reasoning performance, we use two KGQA benchmarks: (a) \dataset{TimeQuestions}~\citep{jia2021complex}, a multi-hop factoid QA dataset over Wikidata requiring answers grounded in specific time periods; and (b) \dataset{MultiTQ}~\citep{chen2023multi}, a large temporal KGQA dataset consisting of 500K unique question-answer pairs, requiring reasoning over multi-hop temporal facts, with varying temporal granularities (day, month, year), and often involves multiple constraints in the question. 5,000 samples are randomly selected from it and divided into 5 test splits. 
(2) \textbf{End-to-End QA with KG Evolution:} We use two domains from the CRAG benchmark~\citep{yang2024crag}: 
(a) \dataset{Movie}, based on IMDB, with questions about films, actors, and awards; and
(b) \dataset{Sports}, covering basketball and soccer matches across major leagues.
Each CRAG question is paired with ground-truth answers, a web document corpus, and a mock domain-specific KG. To simulate real-world KG incompleteness, we downsample key entity types (e.g., only 60\% of \texttt{Movie} and \texttt{Person} nodes are retained in the \dataset{Movie} dataset), requiring recovery through document-driven KG updates.

\textbf{Baselines.} We use five state-of-the-art LLMs. We compare our approach against eight baselines grouped into three categories: (1) \textbf{LLM-only methods}, including IO Prompt (IO), Chain-of-Thought (CoT)~\citep{wei2022chain}, Self-Consistency (SC)~\citep{wangself}, and Retrieval-Augmented Generation (RAG)~\citep{lewis2020retrieval}; (2) \textbf{KG-enhanced methods}, including 1-hop KG~\citep{yang2024crag} (augmenting the LLM with facts from 1-hop KG neighbors of the topic entities) and RAG + 1-hop KG~\citep{yang2024crag}; and (3) \textbf{Multi-hop graph reasoning methods}, including state-of-the-art works: Think-on-Graph~\citep{sunthink} and Plan-on-Graph~\citep{chenplan}. We report the average and standard deviation over 5 runs for all models and baselines.

\begin{table*}[!t]
\vspace{-2em}
\caption{Temporal KGQA results. RAG is not applicable since the datasets don't contain documents.}
\label{table:temporal_results}
\vspace{-0.75em}
\begin{center}
\resizebox{1.0\textwidth}{!}{%
\begin{tabularx}{1.45\textwidth}{@{}c@{}l@{\hspace{0.6em}}c@{\hspace{0.5em}}c@{\hspace{0.5em}}c@{\hspace{0.5em}}c@{\hspace{0.5em}}c@{\hspace{0.5em}}c@{\hspace{0.5em}}c@{\hspace{0.5em}}c@{\hspace{0.5em}}c@{\hspace{0.5em}}c@{\hspace{0.5em}}c@{}}
\toprule
& \thead{\normalsize Datasets $\rightarrow$} & \multicolumn{5}{c}{\dataset{TimeQuestions}} &  \multicolumn{5}{c}{\dataset{MultiTQ}}\\
\cmidrule(lr){3-7} \cmidrule(lr){8-12}
& \thead{\normalsize Methods$\downarrow$}
& \thead{Deep-Seek V3\\671B} & \thead{Qwen 2.5\\72B} & \thead{LLaMA 3.3\\70B} & \thead{GPT-4o\\mini} & \thead{LLaMA 3.1\\8B} 
& \thead{Deep-Seek V3\\671B} & \thead{Qwen 2.5\\72B} & \thead{LLaMA 3.3\\70B} & \thead{GPT-4o\\mini} & \thead{LLaMA 3.1\\8B} 
\\
\midrule
\multirow{4}{*}{\rotatebox{90}{\thead{LLM-Only}}}
& \thead{\normalsize IO}
& \res{54.36}{0.57} & \res{36.31}{0.16} & \res{48.56}{0.37} & \res{35.73}{0.28} & \res{28.04}{0.26} 
& \res{5.20}{0.90} & \res{0.50}{0.17} & \res{4.20}{0.46} & \res{1.37}{0.45} & \res{2.92}{0.48} 
\\
& \thead{\normalsize CoT}
& \res{50.55}{0.54} & \res{44.84}{0.27} & \res{53.41}{0.08} & \res{41.99}{0.11} & \res{38.68}{0.21} 
& \res{2.83}{0.40} & \res{0.94}{0.42} & \res{4.22}{0.71} & \res{5.10}{0.20} & \res{6.42}{0.56} 
\\
& \thead{\normalsize SC}
& \res{49.49}{2.23} & \res{44.30}{0.15} & \res{53.80}{0.20} & \res{42.16}{0.30} & \res{39.98}{0.33} 
& \res{2.43}{0.71} & \res{0.76}{0.29} & \res{4.34}{0.40} & \res{5.03}{0.40} & \res{6.12}{0.90} 
\\
\midrule
\multirow{6}{*}{\rotatebox{90}{\thead{KG-Based}}}
& \thead{\normalsize 1-hop KG}
& \res{37.25}{1.05} & \res{26.97}{.31} & \res{42.27}{0.09} & \res{31.55}{0.16} & \res{33.75}{0.27} 
& \res{7.57}{0.95} & \res{7.14}{1.22} & \res{8.28}{0.99} & \res{5.87}{1.16} & \res{5.06}{0.35} 
\\
& \thead{\normalsize Think-on-Graph}
& \res{32.83}{2.24} & \res{33.07}{0.23} & \res{48.89}{0.27} & \res{32.19}{0.59} & \res{16.50}{0.52} 
& \res{28.90}{3.66} & \res{23.35}{1.76} & \res{25.86}{0.75} & \res{19.57}{2.80} & \res{3.90}{0.41} 
\\
& \thead{\normalsize Plan-on-Graph}
& \res{35.16}{0.91} & \res{32.28}{0.11} & \res{48.58}{0.16} & \res{32.27}{0.26} & \res{27.58}{0.15} 
& \res{29.90}{2.81} & \res{25.16}{0.80} & \res{28.08}{1.04} & \res{19.77}{3.69} & \res{11.04}{0.58} 
\\
& \thead{\normalsize \reasoner{}}
& \fir{67.25}{1.90} & \fir{67.76}{0.28} & \fir{68.63}{0.27} & \fir{65.42}{0.20} & \fir{56.13}{0.12} 
& \fir{39.07}{2.26} & \fir{39.45}{0.38} & \fir{35.30}{1.18} & \fir{34.03}{2.27} & \fir{29.15}{2.15} 
\\
\bottomrule
\end{tabularx}
}
\end{center}
\vspace{-1em}
\end{table*}

\begin{table*}[!t]
\vspace{-0.5em}
\caption{End-to-end experiment results across LLMs of different scales.}
\label{table:end_to_end_results}
\vspace{-0.75em}
\begin{center}
\resizebox{1.0\textwidth}{!}{%
\begin{tabularx}{1.45\textwidth}{@{}c@{}l@{\hspace{0.6em}}c@{\hspace{0.5em}}c@{\hspace{0.5em}}c@{\hspace{0.5em}}c@{\hspace{0.5em}}c@{\hspace{0.5em}}c@{\hspace{0.5em}}c@{\hspace{0.5em}}c@{\hspace{0.5em}}c@{\hspace{0.5em}}c@{\hspace{0.5em}}c@{}}
\toprule
& \thead{\normalsize Datasets $\rightarrow$} & \multicolumn{5}{c}{\dataset{Movie}} &  \multicolumn{5}{c}{\dataset{Sports}}\\
\cmidrule(lr){3-7} \cmidrule(lr){8-12}
& \thead{\normalsize Methods$\downarrow$}
& \thead{Deep-Seek V3\\671B} & \thead{Qwen 2.5\\72B} & \thead{LLaMA 3.3\\70B} & \thead{GPT-4o\\mini} & \thead{LLaMA 3.1\\8B} 
& \thead{Deep-Seek V3\\671B} & \thead{Qwen 2.5\\72B} & \thead{LLaMA 3.3\\70B} & \thead{GPT-4o\\mini} & \thead{LLaMA 3.1\\8B} 
\\
\midrule
\multirow{6}{*}{\rotatebox{90}{\thead{LLM-Only}}}
& \thead{\normalsize IO}
& \res{56.22}{0.10} & \res{34.27}{0.32} & \res{51.75}{0.30} & \res{38.76}{0.31} & \res{34.16}{0.31} 
& \res{38.25}{0.36} & \res{23.78}{0.20} & \res{28.55}{0.56} & \res{26.64}{0.23} & \res{18.60}{0.27} 
\\
& \thead{\normalsize CoT}
& \res{56.52}{0.37} & \res{37.27}{0.43} & \res{59.75}{0.52} & \res{45.37}{0.20} & \res{40.46}{0.59} 
& \res{44.63}{0.62} & \res{29.30}{0.13} & \res{32.15}{0.61} & \res{32.01}{0.23} & \res{26.96}{0.79} 
\\
& \thead{\normalsize SC}
& \res{57.05}{0.37} & \res{37.52}{0.63} & \res{59.15}{0.54} & \res{45.49}{0.47} & \res{43.61}{1.99} 
& \res{45.48}{0.94} & \res{29.44}{0.44} & \res{33.27}{0.51} & \res{32.09}{0.36} & \res{26.49}{0.79} 
\\
& \thead{\normalsize RAG}
& \res{32.39}{0.61} & \res{29.13}{0.27} & \res{30.94}{0.44} & \res{27.55}{0.20} & \res{27.04}{0.38} 
& \res{39.02}{0.62} & \res{34.53}{0.38} & \res{34.91}{0.27} & \res{33.02}{0.14} & \res{29.35}{0.31} 
\\
\midrule
\multirow{7}{*}{\rotatebox{90}{\thead{Original KG}}}
& \thead{\normalsize 1-hop KG}
& \res{25.13}{0.35} & \res{15.75}{0.13} & \res{16.35}{0.10} & \res{18.29}{0.20} & \res{17.59}{0.20} 
& \res{32.71}{1.24} & \res{12.24}{0.13} & \res{16.21}{0.59} & \res{15.65}{0.40} & \res{10.49}{0.12} 
\\
& \thead{RAG + 1-hop KG}
& \res{33.10}{1.06} & \res{31.86}{0.13} & \res{31.08}{0.65} & \res{28.50}{0.31} & \res{27.54}{0.30} 
& \res{36.92}{1.21} & \res{33.78}{0.21} & \res{32.99}{0.51} & \res{30.37}{0.23} & \res{28.46}{0.35} 
\\
& \thead{\normalsize Think-on-Graph}
& \res{23.42}{0.51} & \res{20.53}{1.02} & \res{33.98}{0.99} & \res{20.83}{0.37} & \res{15.36}{0.93} 
& \res{9.89}{0.36} & \res{8.69}{0.51} & \res{11.73}{1.16} & \res{9.42}{0.67} & \res{2.20}{0.27} 
\\
& \thead{\normalsize Plan-on-Graph}
& \res{21.83}{0.45} & \res{15.19}{0.49} & \res{32.71}{0.63} & \res{12.86}{0.27} & \res{17.88}{0.48} 
& \res{15.26}{0.94} & \res{8.27}{0.46} & \res{15.65}{0.87} & \res{12.38}{0.47} & \res{4.63}{0.35} 
\\
& \thead{\normalsize \reasoner{}}
& \res{61.77}{1.45} & \res{50.55}{0.51} & \res{58.90}{1.90} & \res{52.45}{1.44} & \res{45.80}{0.29} 
& \res{53.79}{0.59} & \res{40.14}{0.63} & \res{42.48}{0.98} & \res{44.39}{0.84} & \res{33.36}{1.08} 
\\
\midrule
\multirow{7}{*}{\rotatebox{90}{\thead{Evolved KG}}}
& \thead{\normalsize 1-hop KG}
& \res{21.47}{1.27} & \res{20.18}{0.12} & \res{23.36}{0.22} & \res{20.18}{0.18} & \res{22.51}{0.26} 
& \res{13.40}{3.00} & \res{5.75}{0.13} & \res{7.66}{0.20} & \res{8.96}{0.27} & \res{4.86}{0.26} 
\\
& \thead{RAG + 1-hop KG}
& \res{34.28}{0.51} & \res{33.91}{0.20} & \res{34.97}{0.27} & \res{28.97}{0.20} & \res{30.80}{0.18} 
& \res{31.62}{1.10} & \res{29.63}{0.10} & \res{32.80}{0.27} & \res{28.50}{0.40} & \res{26.45}{0.30} 
\\
& \thead{\normalsize Think-on-Graph}
& \res{25.09}{1.17} & \res{31.82}{0.34} & \res{40.21}{0.51} & \res{24.60}{0.18} & \res{23.65}{1.37} 
& \res{24.92}{2.86} & \res{23.55}{0.56} & \res{22.29}{0.27} & \res{15.81}{1.41} & \res{10.28}{1.10} 
\\
& \thead{\normalsize Plan-on-Graph}
& \res{23.78}{0.45} & \res{31.19}{0.40} & \res{41.17}{0.46} & \res{24.54}{0.54} & \res{24.71}{0.87} 
& \res{28.04}{1.98} & \res{23.46}{0.51} & \res{28.08}{0.85} & \res{16.04}{1.05} & \res{13.46}{0.97} 
\\
& \thead{\normalsize \reasoner{}}
& \fir{63.13}{0.10} & \fir{57.84}{0.70} & \fir{64.85}{0.71} & \fir{57.52}{0.53} & \fir{51.11}{0.79} 
& \fir{55.61}{0.33} & \fir{47.90}{0.23} & \fir{51.17}{0.62} & \fir{51.64}{1.98} & \fir{37.06}{1.58} 
\\
\bottomrule
\end{tabularx}
}
\end{center}
\vspace{-2em}
\end{table*}

\vspace{-0.25em}
\subsection{Experimental Results}

\textbf{A1: Comparative Results on Temporal Reasoning.}
Table~\ref{table:temporal_results} summarizes our results on temporal QA benchmarks. Across all datasets and model scales, \reasoner{} consistently outperforms strong baselines. It achieves up to \textbf{23.3\%} and \textbf{18.1\%} absolute improvement over the best baseline in \dataset{TimeQuestions} and \dataset{MultiTQ}, respectively. Moreover, the performance improves further with larger LLMs, indicating that more capable models better follow \reasoner{}’s decomposition strategy, rank relevant relations, and explore local graph neighborhoods. Interestingly, although Qwen 2.5 and LLaMA 3.3 are similar in size, the accuracy of their LLM-only methods diverges, potentially due to differences in pretraining of these models. However, once grounded in the KG through \reasoner{}, their performance converges. This suggests that \reasoner{} mitigates discrepancies in internal knowledge by anchoring inference in external, structured signals.

On \dataset{TimeQuestions}, \reasoner{} improves absolute accuracy by 15--23\% compared to the best LLM-only baseline, and by 19--33\% over the best KG-enhanced and graph reasoning baselines. For example, on Qwen 2.5–72B, our method improves accuracy from 44.84\% (best LLM-only) to 67.76\% (+22.9\%) and from 33.07\% (best KG-based method) to 67.76\% (+34.7\%). Similarly, on GPT-4o-mini, \reasoner{} lifts accuracy from 42.16\% (LLM-only) and 32.27\% (KG-based) to 65.42\%, yielding relative gains of +23.3\% and +33.2\%, respectively.

On the more challenging \dataset{MultiTQ} dataset, \reasoner{} maintains its lead despite lower overall accuracy due to higher temporal ambiguity. It improves accuracy by 23--38\% compared to the best LLM-only baseline, and outperforms the best KG-enhanced and reasoning baselines by 8--18\%.

\textbf{A2: \updater{} Boosts Reasoning Performance.}
Table~\ref{table:end_to_end_results} presents QA accuracy before and after applying \updater{}. All multi-hop graph reasoning models, including \reasoner{}, PoG, and ToG, benefit from the KG update, with accuracy gains of 2--8\%. In contrast, 1-hop KG and RAG+1-hop KG baselines show limited or even negative gains, especially on \dataset{Sports}. This is due to their reliance on indiscriminate context concatenation: 1-hop KG retrieves all neighboring facts until the context window is full. In domains like sports, where a single team can accumulate hundreds of historical matches over multiple seasons, this dense retrieval overwhelms the model and introduces irrelevant information, ultimately degrading performance.

\reasoner{} not only outperforms all baselines on the original KG—as shown in the temporal QA task—but also benefits significantly from KG updates. In both domains, its performance increases by 2–8\% after applying \updater{}, validating the quality of the constructed facts. For example, in the \dataset{Movie} dataset with GPT-4o-mini, \reasoner{} improves accuracy from 52.45\% (original KG) to 57.52\% (updated KG), a +5\% boost. Moreover, with the help of \updater{}, a small LLaMA 3.1–8B model trained in December 2023 achieves a 18.6 → 37.0\% (+18.4\%) accuracy gain, which is comparable to directly prompting a 671B DeepSeek-V3 model (38.3\%) trained in July 2024. This demonstrates that our framework enables strong QA performance even for smaller models by evolving and grounding the KG appropriately, and closes the gap between the small and large models.

\subsection{Ablation Studies on \reasoner{}}
We conduct ablation studies on \dataset{TimeQuestions} and \dataset{Sports}, which contains knowledge up to Mar 2024, to assess the contribution of each component in \reasoner{}. All results are reported using LLaMA 3.3 (70B) and LLaMA 3.1 (8B), both with knowledge cutoff in Dec 2023, to isolate the effect of reasoning over knowledge in KGs.

\begin{wraptable}{r}{0.5\columnwidth}
\vspace{-2em}
\captionof{table}{Ablation studies on \reasoner{}.}
\label{table:ablation}
\vspace{-0.75em}
\begin{center}
\centering
\resizebox{0.5\textwidth}{!}{%
\begin{tabularx}{0.74\textwidth}{@{}c@{}l@{\hspace{0.8em}}c@{\hspace{0.6em}}c@{\hspace{0.6em}}c@{\hspace{0.6em}}c@{\hspace{0.6em}}c@{\hspace{0.6em}}c@{\hspace{0.6em}}c@{\hspace{0.6em}}c@{\hspace{0.6em}}c@{\hspace{0.6em}}c@{\hspace{0.6em}}c@{}}
\toprule
& \thead{\normalsize Datasets $\rightarrow$} & \multicolumn{2}{c}{\dataset{TimeQuestions}} &  \multicolumn{2}{c}{\dataset{Sports}}\\
\cmidrule(lr){3-4} \cmidrule(lr){5-6}
& \thead{\normalsize Methods$\downarrow$}
& \thead{LLaMa 3.3\\70B} & \thead{LLaMA 3.1\\8B} 
& \thead{LLaMa 3.3\\70B} & \thead{LLaMA 3.1\\8B} 
\\
\midrule
& \thead{\normalsize \reasoner{}}
& \fir{68.63}{0.27} & \fir{56.13}{0.12}
& \fir{51.17}{0.62} & \fir{37.06}{1.58}
\\
& \thead{ w/o Route Decomposition}
& \res{67.60}{0.55} & \res{43.98}{0.61}
& \res{47.90}{0.76} & \res{26.52}{1.27}
\\
& \thead{ w/o Global Search}
& \res{64.12}{0.30} & \res{52.35}{0.79}
& \res{42.46}{0.84} & \res{32.67}{1.62}
\\
\bottomrule
\end{tabularx}
}
\end{center}
\centering
\includegraphics[width=0.8\linewidth]{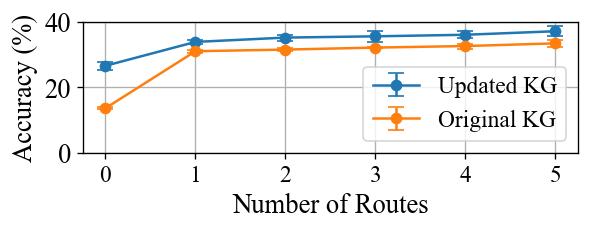}
\captionof{figure}{Route-wise performance in \dataset{Sports} on the LLaMA 3.1 8B model.}
\label{fig:route_ablation}
\vspace{-1em}
\end{wraptable}
\textbf{Impact of Multi-Route Decomposition and Global Search.}
Table~\ref{table:ablation} shows that removing either route decomposition or global search leads to significant accuracy drops.
(1) Removing multi-route decomposition causes the largest degradation (up to 20\%), confirming the importance of soft reasoning guidance for reasoning over KGs.
(2) Without global search, performance drops by 4–9\%, showing the need for accurate entity grounding.

\textbf{Diverse Routes Improve Accuracy.}
Figure~\ref{fig:route_ablation} reports route-wise QA performance on \dataset{Sports}. Accuracy improves steadily with more routes, validating that route diversity enhances KG coverage and reasoning robustness. 

These results underscore the value of both flexible multi-path reasoning and temporal grounding in achieving strong QA performance.

%% file: sections/6_related_works.tex
\noindent\textbf{KG Reasoning.}
Reasoning over knowledge graphs (KGs) involves drawing structured inferences to support tasks like question answering and decision-making~\citep{pan2024unifying}. Existing methods typically fall into two categories: (1) embedding-based approaches and (2) LLM-driven approaches.
Embedding-based methods encode entities and relations into dense vectors to retrieve relevant subgraphs. Recent work such as SR~\citep{zhang2022subgraph}, UniKGQA~\citep{jiang2023unikgqa}, and ReasoningLM~\citep{jiang2023reasoninglm} adopt this strategy, but may struggle with noise and limited interpretability.
LLM-driven approaches leverage large language models to guide or perform reasoning. StructGPT~\citep{jiang2023structgpt} generates structured queries, ToG~\citep{sunthink} identifies salient subgraphs, and PoG~\citep{chenplan} decomposes queries into multi-step plans. KG-Agent~\citep{jiang2024kg} extends this with tool-augmented traversal. However, most approaches assume a static KG and treat temporal attributes as metadata, limiting their ability to handle evolving knowledge.

\noindent\textbf{KG Extraction.}
Traditional KG construction techniques often rely on rule-based extraction from curated resources (e.g., YAGO~\citep{suchanek2007yago}), but these approaches are brittle and hard to scale.
More recent efforts turn to LLMs for end-to-end KG extraction from raw text. Techniques such as JointIE~\citep{qiao2022joint}, EXTRACT~\citep{zhang2024extract}, and UP-KG~\citep{hatem2024up} formulate KG extraction as a text-to-graph generation task. These techniques demonstrate promising performance but often overwrite existing knowledge without accounting for temporal drift or contradictions.
Our work builds on this line of work but introduces a temporal-aware, contradiction-tolerant evolution process, allowing the KG to store multiple values, track confidence, and maintain historical facts over time.

%% file: sections/7_conclusion.tex
We present a unified framework for temporal reasoning over evolving knowledge graphs, that combines a multi-hop reasoning module (\reasoner{}) with a temporal-aware KG evolution method (\updater{}). By jointly addressing contradiction resolution and temporal trend modeling, our method enables accurate and robust reasoning in dynamic knowledge environments. Comprehensive experiments on temporal QA and end-to-end KG update benchmarks show substantial improvements over LLM-only and KG-enhanced baselines. Although our framework demonstrates strong performance, it assumes access to clean temporal cues and may face scalability challenges on very large corpora. Addressing these limitations is an interesting direction for future work.

%% file: sections/appendix.tex
\section{Broader Impacts} \label{sec:impacts}
This work contributes to advancing temporal reasoning over evolving knowledge graphs, with potential positive applications in areas such as personal assistants, scientific discovery, legal research, and financial intelligence. By enabling more accurate and context-aware information retrieval, our framework could improve decision-making in dynamic, real-world environments.

However, several risks must also be considered. First, if the KG update mechanism integrates adversarial or biased sources, the system could amplify misinformation. Second, overreliance on automatically updated knowledge bases could reduce human oversight in high-stakes domains. Finally, evolving KGs might be misused for surveillance or automated profiling if applied to personal or sensitive data without consent. Mitigating these risks will require careful data curation, source verification, and transparency in how facts are scored, selected, and surfaced to end users.

\section{KG Update Details}\label{sec:update_details}
\paragraph{Edge Insertion and Update:}
\begin{itemize}[topsep=1pt, leftmargin=15pt]
\item No matching relation between $v_s$ and $v_t$:
\[
\mathcal{R}_{\text{existing}} = \emptyset \quad \Rightarrow \quad \textbf{Insert } \hat{e}
\]
\item Exact same relation exists, no new info:
\[
\exists e’ = (v_s, r^*, v_t) \in \mathcal{E}_t \text{ and } \pi(\hat{e}) \subseteq \pi(e’) \quad \Rightarrow \quad \textbf{Skip}
\]
\item Exact same relation exists, new info:
\[
\exists e’ = (v_s, r^*, v_t) \in \mathcal{E}_t \text{ and } \pi(\hat{e}) \not\subseteq \pi(e’) \quad \Rightarrow \quad \textbf{Merge}
\]
\item Synonym relation exists with same $(v_s, v_t)$:
\[
\exists r’ \in \mathcal{R}_{\text{existing}},\ r’ \sim r^,\ r’ \ne r \quad \Rightarrow \quad \textbf{Map to } r’ \text{ and } \textbf{Merge}
\]
\item Synonym relation exists with different edges:
\[
\exists r’ \sim r^* \text{ but } (v_s, r’, v_t) \notin \mathcal{E}_t \quad \Rightarrow \quad \textbf{Map to } r’ \text{ and } \textbf{Insert}
\]
\end{itemize}

\section{Additional Experiment Details}
\paragraph{Models.} We evaluate our methods using five state-of-the-art LLMs, ordered by parameter size: DeepSeek V3 (671B) \citep{liu2024deepseek}, Qwen 2.5 (72B) \citep{qwen2.5}, LLaMA 3.3 (70B) \citep{grattafiori2024llama}, GPT-4o-mini \citep{openai2024gpt4omini}, and LLaMA 3.1 (8B) \citep{grattafiori2024llama}. All models follow the same QA protocol for fair comparison. We use a standard lightweight sentence-transformers model (IBM/slate-125m-english-rtrvr-v2~\citep{ibm2024slate125m}) to generate semantic embeddings. We performed the KG updating using the LLaMA 3.3 (70B) model. All the models are accessible through online API providers, such as OpenAI, DeepSeek, and OpenRouter.

\subsection{Licenses for Existing Assets} \label{sec:license}
We use the following models and assets in this work, all of which are properly cited and used under their respective licenses:
\begin{itemize}
    \item \textbf{DeepSeek V3 (671B)}: Accessed via DeepSeek API, under the DeepSeek open-source model license available at \url{https://github.com/deepseek-ai/DeepSeek-LLM/blob/main/LICENSE-MODEL}.
    \item \textbf{Qwen 2.5 (72B)}: Accessed via OpenRouter, based on Qwen model family released by Alibaba under a commercial-use friendly license.
    \item \textbf{LLaMA 3.3 \& 3.1 (70B, 8B)}: Accessed via OpenRouter API, follows Meta’s LLaMA 3 license available at \url{https://ai.meta.com/llama/license/}.
    \item \textbf{GPT-4o-mini}: Accessed through OpenAI API, used in accordance with the terms of service and usage policies listed at \url{https://openai.com/policies/terms-of-use}.
    \item \textbf{IBM Slate-125M-english-rtrvr-v2:} Released by IBM under Apache 2.0 License via \url{https://huggingface.co/ibm-granite/granite-embedding-125m-english}.
\end{itemize}
All datasets used in the paper (TimeQuestions, MultiTQ, CRAG benchmark) are publicly released by their authors for research purposes and properly cited in the paper.

\paragraph{Compute Resources.} \label{sec:compute}
All experiments involving large language models were conducted via public API providers. Specifically, we accessed models through the OpenAI and OpenRouter platforms, which internally manage GPU compute infrastructure. As such, we do not control the underlying hardware but follow their standard rate limits and access conditions.

All preprocessing, knowledge graph construction, and hosting were performed on a cloud server equipped with an Intel Xeon E5-2698 CPU (40 cores, 2.2 GHz) and 500GB of main memory. Each full run of a QA benchmark (including reasoning) takes less than 2–3 hours, depending on model size and API latency. The KG update takes longer, which requires processing more than 3,400 corpora with an average length of 10,000 words, and usually can be accomplished within 24 hours. No dedicated GPU resources were used beyond the API endpoints.

\section{Prompt Templates} \label{sec:prompt_template}
\newtcolorbox{promptbox}[2][]{
  breakable,
  enhanced,
  colback=#2!5,
  colframe=#2!60!black,
  colbacktitle=#2!20,
  coltitle=black,
  fonttitle=\bfseries,
  title=#1,
  sharp corners=south,
  rounded corners=north,
  attach boxed title to top center={
    yshift=-1mm,
    yshifttext=-1mm
  },
  boxed title style={
    size=normal,
    colframe=#2!60,
    boxrule=0pt
  }
}

\begin{promptbox}[Prompt Templates for Route Planning]{cyan}
You are a helpful assistant who is good at answering questions in the {domain} domain by using knowledge from an external knowledge graph. Before answering the question, you need to break down the question so that you may look for the information from the knowledge graph in a step-wise operation. Hence, please break down the process of answering the question into as few sub-objectives as possible based on semantic analysis. A query time is also provided; please consider including the time information when applicable.\\
    
There can be multiple possible route to break down the question, aim for generating {route} possible routes. Note that every route may have a different solving efficiency, order the route by their solving efficiency.\\
Return your reasoning and sub-objectives as multiple lists of strings in a flat JSON of format: {{"reason": "...", "routes": [[<a list of sub-objectives>], [<a list of sub-objectives>], ...]}}. (TIP: You will need to escape any double quotes in the string to make the JSON valid)\\
\\
-Example-\\
Q: Which of the countries in the Caribbean has the smallest country calling code?\\
Query Time: 03/05/2024, 23:35:21 PT\\
Output: \{
"reason": "The most efficient route involves directly identifying Caribbean countries and their respective calling codes, as this limits the scope of the search. In contrast, routes that involve broader searches, such as listing all country calling codes worldwide before filtering, are less efficient due to the larger dataset that needs to be processed. Therefore, routes are ordered based on the specificity of the initial search and the subsequent steps required to narrow down to the answer.",\\
"routes": [["List all Caribbean countries", "Determine the country calling code for each country", "Identify the country with the smallest calling code"],
["Identify Caribbean countries", "Retrieve their country calling codes", "Compare to find the smallest"],
["Identify the smallest country calling code globally", "Filter by Caribbean countries", "Select the smallest among them"],
["List all country calling codes worldwide", "Filter the calling codes by Caribbean countries", "Find the smallest one"]]
\}\\
\\
Q: <query>\\
Query Time: <query time>\\
Output Format (flat JSON): \{\{"reason": "...", "routes": [[<a list of sub-objectives>], [<a list of sub-objectives>], ...]\}\}\\
Output:
\end{promptbox}

\begin{promptbox}[Prompt Templates for Context-Aware Global Initialization]{green}
-Goal-\\
You are presented with a question in the {domain} domain, its query time, and a potential route to solve it.\\
\\
1) Determine the topic entities asked in the query and each step in the solving route. The topic entities will be used as source entities to search through a knowledge graph for answers.
It's preferrable to mention the entity type explictly to ensure a more precise search hit.\\
\\
2) Extract those topic entities from the query into a string list in the format of ["entity1", "entity2", ...].\\
Consider extracting the entities in an informative way, combining adjectives or surrounding information. \\
A query time is provided - please consider including the time information when applicable.\\
\\
*NEVER include ANY EXPLANATION or NOTE in the output, ONLY OUTPUT JSON*  \\
\\
-Examples-\\
<few-shot examples>\\
\\
Question: <query>\\
Query Time: <query time>\\
Solving Route: <route>\\
Output Format: ["entity1", "entity2", ...]\\
Output:
\end{promptbox}

\begin{promptbox}[Prompt Templates for Global Initialization Relevance Scoring]{red}
-Goal-\\
You are presented with a question in the {domain} domain, its query time, a potential route to solve it, and a list of entities extracted from a noisy knowledge graph.\\
The goal is to identify all possible relevant entities to answering the steps in the solving route and, therefore, answer the question.\\
You need to consider that the knowledge graph may be noisy and relations may split into similar entities, so it's essential to identify all relevant entities.\\
The entities' relevance would be scored on a scale from 0 to 1 (use at most 3 decimal places, and remove trailing zeros; the sum of the scores of all entities is 1). \\
\\
-Steps-\\
1. You are provided a set of entities (type, name, description, and potential properties) globally searched from a knowledge graph that most similar to the question description, but may not directly relevant to the question itself.\\
Given in the format of "ent\_i: (<entity type>: <entity name>, desc: "description", props: \{key: [val\_1 (70\%, ctx:"context"), val\_2 (30\%, ctx:"context")], ...\})"\\
where "i" is the index, the percentage is confidence score, "ctx" is an optional context under which the value is valid. Each property may have only a single value, or multiple valid values of vary confidence under different context.\\
\\
2. Score *ALL POSSIBLE* entities that are relevant to answering the steps in the solving route and therefore answering the question, and provide a short reason for your scoring.\\
Return its index (ent\_i) and score into a valid JSON of the format: \{"reason": "reason", "relevant entities": \{"ent\_i": 0.6, "ent\_j": 0.3, ...\}\}. (TIP: You will need to escape any double quotes in the string to make the JSON valid)\\
\\
*NEVER include ANY EXPLANATION or NOTE in the output, ONLY OUTPUT JSON*  \\
\\
-Examples-\\
<few-shot examples>\\
\\
Question: <query>\\
Query Time: <query time>\\
Solving Route: <route>\\
Entities: <topk entities str>\\
\\
Output Format (flat JSON): \{"reason": "reason", "relevant\_entities": \{"ent\_i": 0.6, "ent\_j": 0.3, ...\}\}\\
Output:
\end{promptbox}